\documentclass[10pt,twocolumn,letterpaper]{article}

\usepackage{cvpr}              % To produce the CAMERA-READY version
\usepackage{rotating}
\usepackage{algorithm,algpseudocode}
\usepackage{algorithmicx}
\usepackage[dvipsnames]{xcolor}

\definecolor{cvprblue}{rgb}{0.21,0.49,0.74}
\definecolor{forestgreen}{rgb}{0.13, 0.55, 0.13}
\usepackage[pagebackref,breaklinks,colorlinks,allcolors=cvprblue]{hyperref}
\usepackage{soul}
\usepackage{float}
\usepackage{graphicx}

\def\paperID{5759} % *** Enter the Paper ID here
\def\confName{CVPR}
\def\confYear{2025}

\title{InTraGen: Trajectory-controlled Video Generation for Object Interactions}
\begin{document}

\author{Zuhao Liu\textsuperscript{1,3*\textdagger} 
\quad
Aleksandar Yanev\textsuperscript{1*\textdagger}
\quad
Ahmad Mahmood\textsuperscript{1,2\textdagger}
\quad
Ivan Nikolov\textsuperscript{4}
\quad
Saman Motamed\textsuperscript{1}\\
\quad
Wei-Shi Zheng\textsuperscript{3}
\quad
Xi Wang\textsuperscript{1,2}
\quad 
Lei Sun\textsuperscript{1}
\quad 
Luc Van Gool\textsuperscript{1,2}
\quad 
Danda Pani Paudel\textsuperscript{1}
\\
    \normalsize\textsuperscript{1} INSAIT, Sofia University “St. Kliment Ohridski”, Bulgaria \quad
    \normalsize\textsuperscript{2} ETH Zurich, Switzerland \\
    \normalsize\textsuperscript{3} Sun Yat-sen University, China \quad
    \normalsize\textsuperscript{4} Aalborg University, Denmark
}

\twocolumn[{%
\renewcommand\twocolumn[1][]{#1}%
\maketitle
\begin{center}
    \centering
    %\captionsetup{type=figure}
    \captionsetup{type=figure}
    \includegraphics[width=0.85\linewidth]{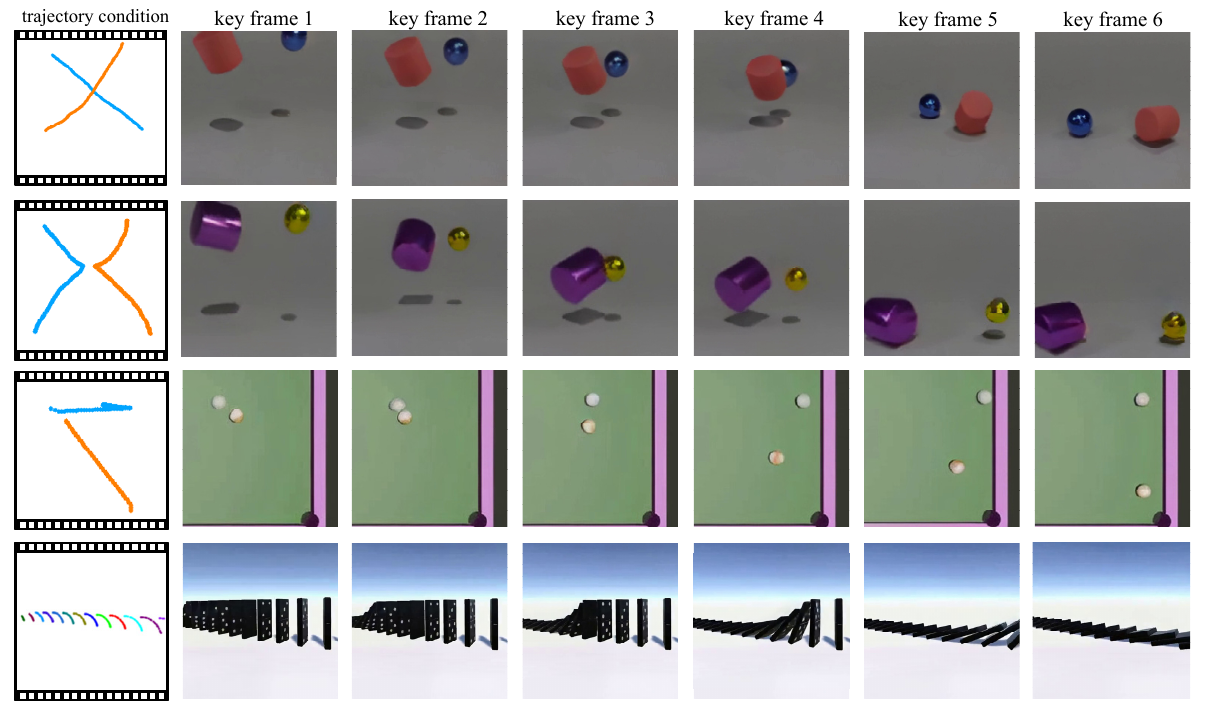}
    \captionof{figure}{Our pipeline generates realistic object interactions—such as crossing, collision, and falling—based on input trajectories. The figure displays four different scenarios. For each scenario, a desired trajectory is provided (left), and the corresponding keyframes (right) illustrate interactions that follow the specified trajectory conditions.}
    \label{first_page_fig}
\end{center}
}]

\def\thefootnote{*}\footnotetext{Authors contributed equally.}
\def\thefootnote{\textdagger}\footnotetext{Work was done during the internship at INSAIT.}

\begin{abstract}
Advances in video generation have significantly improved the realism and quality of created scenes. This has fueled interest in developing intuitive tools that let users leverage video generation as world simulators. Text-to-video (T2V) generation is one such approach, enabling video creation from text descriptions only. Yet, due to the inherent ambiguity in texts and the limited temporal information offered by text prompts, researchers have explored additional control signals like trajectory-guided systems, for more accurate T2V generation. Nonetheless, methods to evaluate whether T2V models can generate realistic interactions between multiple objects are lacking. We introduce InTraGen, a pipeline for improved trajectory-based generation of object interaction scenarios. We propose 4 new datasets and a novel trajectory quality metric to evaluate the performance of the proposed InTraGen. To achieve object interaction, we introduce a multi-modal interaction encoding pipeline with an object ID injection mechanism that enriches object-environment interactions. 
Our results demonstrate improvements in both visual fidelity and quantitative performance.
Code and datasets are available at https://github.com/insait-institute/InTraGen

%Text-to-video (T2V) generation has gained much attention in computer vision due to its large potential in many areas. Due to the ambiguity of text conditions and the lack of temporal understanding, some research explores T2V generation based on additional control signals, especially for trajectory-guided systems, due to its availability and user-friendliness. However, there is a lack of evaluation methods to evaluate whether these controllable t2v methods can generate multi-objects simultaneously with realistic interaction. Therefore, this work proposes a new benchmark (VOIBench) to evaluate the object interaction effect in controllable video generation with four different datasets and a novel interaction quality metric. We propose a trajectory-based DiT architecture to achieve controllable video generation and an ID injection mechanism to improve the object-interaction ability. The proposed methods achieve excellent visual and quantitative results in the proposed VOIBench. 

\end{abstract}
\section{Introduction}
Video generation is a rapidly developing field, with many excellent new works \cite{blattmann2023stable, ma2024latte, wang2023lavie, girdhar2023emu, chen2023videocrafter1, kondratyuk2023videopoet, bartal2024lumiere}. Most video generation methods are currently based on diffusion models, except some decoder-based models such as VideoPoet \cite{kondratyuk2023videopoet}.
Recently, the Diffusion Transformer (DiT) architecture \cite{peebles2023scalable} has gained prominence over the previously popular U-Net architecture, particularly in video generation tasks \cite{ma2024latte, opensora, pku_yuan_lab_and_tuzhan_ai_etc_2024_10948109}, offering enhanced performance and greater flexibility.
Video generation presents greater challenges than image generation due to the complex temporal patterns involved, such as body movements and object interactions. While closed-source models like OpenAI's Sora \cite{sora2024} and Meta's Movie Gen \cite{meta_movie_gen2024} have demonstrated impressive capabilities in limited demos, open-source models continue to face difficulties in producing comparable results. 

\begin{figure*}[t]
    \centering
    \includegraphics[width=\linewidth]{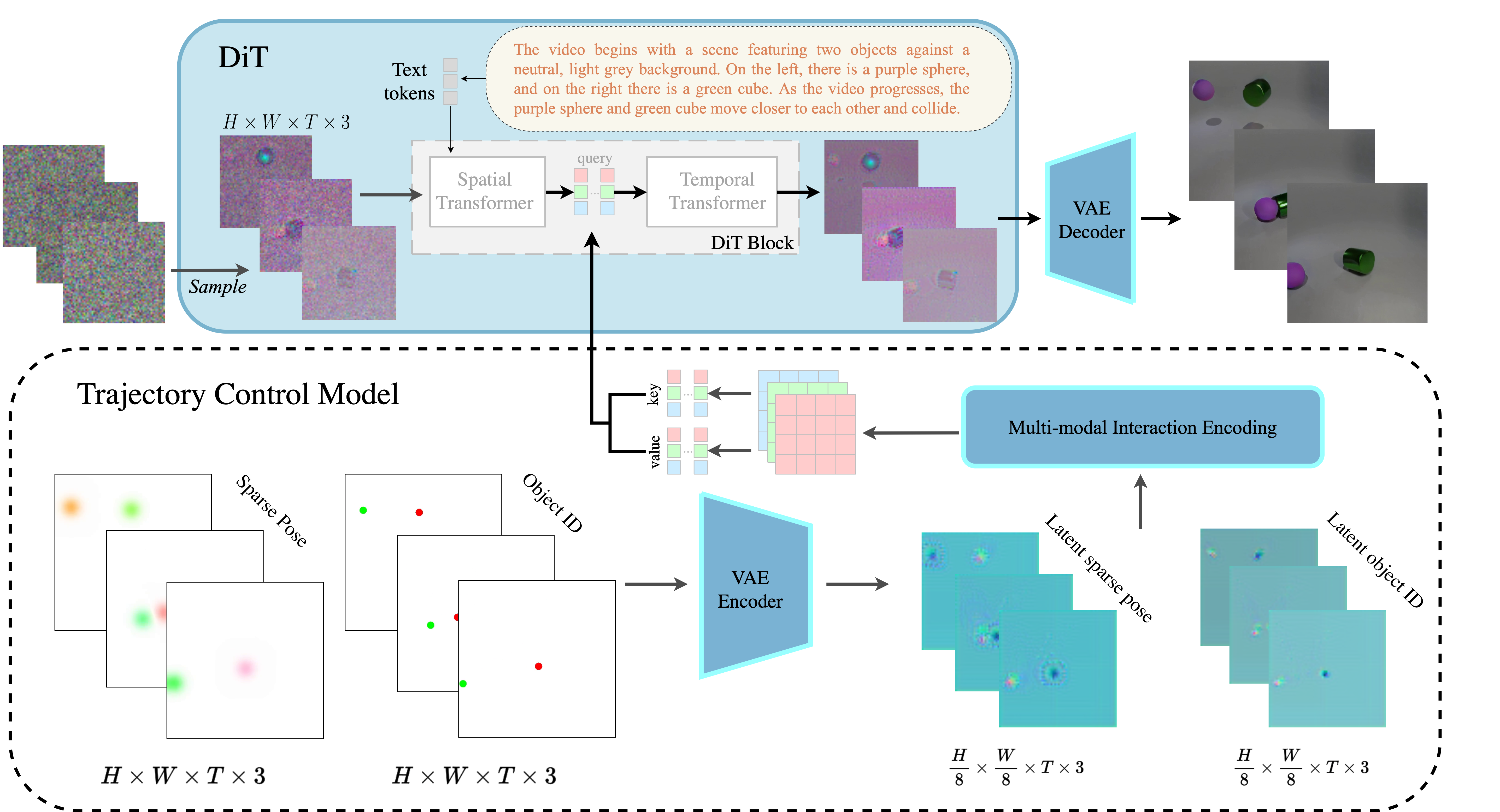}
    \caption{\emph{\textbf{Overview of the Model Architecture:}} The figure illustrates the key components of InTraGen: the Trajectory Control Model and the DiT block. Object IDs and sparse poses are first encoded using a VAE to generate latent representations. The resulting latents are then integrated through the Multi-Modal Interaction Encoding pipeline (Figure \ref{fig:multi-modal_interaction_encoding}) to encode the rich object interaction information.}
    \label{fig:enter-label}
    \vspace{-5mm}
\end{figure*}

Current video generation training pipelines usually generate video captions from large-scale datasets, where the text descriptions can remain ambiguous. Then, a generative model is trained with captions as conditions and videos as outputs, where temporal reasoning deficiency can be observed. These two main challenges hinder the performance of current text-to-video generative models. Text ambiguity happens because video caption models such as Pllava \cite{xu2024pllava} and SharedGPT4Video \cite{chen2024sharegpt4video} fail to adequately cover all the details in a video, especially when the video includes complex object interactions (see Appendix A for more information). Mostly, they can only cover the layout of the video frames but struggle to deal with the object interactions. Temporal reasoning deficiency occurs due to the text encoders that are applied in the generative model like T5 \cite{raffel2020exploring} and CLIP \cite{radford2021learning}. They lack reasoning capabilities, especially regarding the temporal dimension (see Appendix B for more information). The text encoder struggles to understand text prompts that describe temporal information because they are mostly trained on images for the input prompt \cite{Bain21, Xu_2021}. For example, if the input prompt is ``Initially, the red car appears to the left of the frame. As the video progresses, a blue car appears and hits the red car". The video generation model is likely to generate red and blue cars in the first frames, ignoring the temporal sequence of events (please see Appendix B).

Therefore, it is important to apply additional control signals to the video generation model in order to mitigate the influence of the above two challenges. That said, providing any kind of control signal having to do with the time dimension is more challenging. For instance, ControlNet \cite{zhang2023adding} for image generation supports various control modalities, such as `canny edge', `human pose', and `depth map'. However, producing similar control signals for the video domain is more difficult due to the complexities induced by motion. A control signal more suitable for use in the video domain is a set of trajectories, defining the coordinates of different objects in each frame of the video. Trajectories can be extracted from videos using an object tracking model or drawn by users via an interactive interface. When an object tracking model is used, it outputs a list of coordinates with corresponding timestamps for each object. If user-drawn trajectory lines are used, we sample points along the drawn trajectory and distribute them across the duration of the generated video, assigning each point a corresponding timestamp.
% \luc{not so clear what the link with time is if you draw 1 or more trajectories... which point along each trajectory is reached by when ? - Sam: discussing with Zuhao on slack. trajectories are actually videos...have to improve the figure and language here}

%Because trajectories play a crucial role in conveying interaction information for video generation due to their high flexibility, 
Our work focuses on using trajectories to enhance interaction quality. Some recent approaches \cite{zhang2024tora, wang2024motionctrl, yin2023dragnuwa} condition the diffusion model on trajectories represented as sparse optical flow. However, these methods often overlook the presence of both dynamic (moving) and static (stationary) objects, as well as how they precisely interact, making it difficult to accurately associate unique trajectories with the correct objects. Sparse optical flow is limited to capturing dynamic and interactive movements, failing to account for stationary objects or the uniqueness of each object. Our challenge, therefore, is to model dynamic, static, and interactive information while uniquely associating distinct trajectories with individual objects — a requirement which we refer to as interaction-level control.

This paper provides an overall solution to the above challenges. We present InTraGen, a novel pipeline for trajectory-based generation in scenarios where objects interact. Our method introduces a multi-modal interaction encoding mechanism to better capture the dynamic, static, and interactive information. To better evaluate the model performance in interaction scenarios, we use Blender to generate a video dataset with rich object interactions. The dataset contains 50K videos and is divided into four subsets: 1) \textit{Extended-MoVi} 2) \textit{pool game} 3) \textit{dominoes} and 4) \textit{football} videos. Due to the importance of control via trajectories, we have the coordinates (trajectories) for each object in the dataset. The object trajectories are important when training the video generation model and when evaluating its performance in correctly representing object interactions. In addition, we provide a trajectory quality metric to evaluate the performance of the generated videos. 

%We propose a trajectory-based DiT model with a novel ID control injection mechanism to generate videos that follow given trajectories with better interaction performance.

%This paper provides an overall solution to the above challenges. We propose a novel benchmark named Video Object Interaction Benchmark (VOIBench). Under this benchmark, we use Blender to generate a video dataset with rich object interaction. The dataset contains 50K videos and is divided into four sub-sets: general object interaction, pool game, domino, and soccer datasets. Due to the importance of controlling with trajectory, we have the coordinates (trajectories) for each object in the dataset. The object's trajectories are important in training the video generation model and evaluating its performance towards object interactions. In addition,  we provide an interaction quality metric in our benchmark to evaluate the performance of the generated videos regarding object interaction. We propose a trajectory-based DiT model that can generate video following the trajectory. We further propose a novel ID control injection method for better interaction performance. 

In summary, this paper makes the following contributions:
\begin{enumerate}
    \item We propose \textbf{InTraGen}, a novel trajectory-controlled video generation model designed to produce videos with rich object interactions.
    \item We introduce \textbf{four rendered video datasets} with diverse interaction scenarios, along with an \textbf{trajectory quality metric} to evaluate the quality of generated object interactions.
    \item We introduce a \textbf{multi-modal interaction encoding} pipeline that enriches object-environment interactions.
    \item Our results demonstrate substantial improvements in both visual fidelity and quantitative performance for video generation tasks.
\end{enumerate}

%(1) We introduce \textbf{four rendered datasets}, with rich interaction information as well as an interaction quality metric to evaluate the quality of object interaction. (2) 
%(2) We propose a \textbf{trajectory-based DiT architecture} to generate videos that follow the control trajectories. (3) We develop an \textbf{ID control injection mechanism} to improve the object-interaction ability of the model.

% (1) Propose a new benchmark to evaluate the object interaction of the video generation. The benchmark contains four rendered datasets with rich interaction information. The benchmark also contains an interaction quality metric to evaluate the quality of the object interaction.

% (2) Propose a trajectory-based DiT architecture to generate videos following the control trajectories.

% (3) Propose ID control injection mechanism to improve the object-interaction ability.

% (4) Achieve excellent results in the VOIBench with better FID, FVD, and interaction quality compare with SOTA.

\section{Related work}
\label{sec:related_works}
\subsection{Text-to-Video Generation}
The task of text-to-video (T2V) generation aims to produce videos conditioned on text input. In recent years, diffusion models have emerged as the leading approach for T2V generation, replacing previous methods such as Generative Adversarial networks (GANS) \cite{Tulyakov_2018, brooks2022generating, Skorokhodov_2022} and Variational Autoencoders (VAEs)\cite{kingma2022autoencodingvariationalbayes, yan2021videogpt}. Many recent works, like Stable Video Diffusion \cite{blattmann2023stable}, use the U-Net \cite{ronneberger2015unetconvolutionalnetworksbiomedical} architecture as the denoising model, establishing a strong foundation that has inspired numerous subsequent works \cite{chen2023videocrafter1, chen2024videocrafter2, guo2023i2v}.  More recently, some works have started to adopt transformers for denoising \cite{vaswani2023attentionneed, kondratyuk2023videopoet}. Among these works, closed-source T2V models such as Sora \cite{sora2024} and Meta Movie Gen \cite{meta_movie_gen2024}, while achieving excellent results, require massive computing resources to train the model. In contrast, open-source T2V models such as Open-Sora \cite{opensora} and Open-Sora-Plan \cite{pku_yuan_lab_and_tuzhan_ai_etc_2024_10948109} do not achieve the same level of success but perform relatively well, mostly relying on a Diffusion Transformer-based (DiT) video generation architecture named Latte \cite{ma2024latte}. 
However, a key limitation of current T2V methods is their lack of precise control over the generated content, in particular over the way objects move. Such lack of control has motivated 'controllable video generation'.
% \xw{I would also write one sentence to point out the limitation of the T2V methods, which ideally is also the motivation for developing controllable video generation methods. } 
% \luc{how about: ... precise control over the generated content, in particular over the way objects move. Such lack of control has motivated 'controllable video generation'.}

\subsection{Controllable Video Generation}
Controllable video generation has recently garnered substantial attention, with the aim of producing videos that adhere to user-defined trajectories \cite{hu2023videocontrolnetmotionguidedvideotovideotranslation, zhang2024tora, Motamed_2024_CVPR, wang2024motionctrl, peng2024controlnextpowerfulefficientcontrol}. Traditional approaches to video generation faced challenges in maintaining temporal consistency and handling complex motion control. With the advent of diffusion models, significant strides have been made in producing high-quality and consistent video content, enabling new frameworks for precise motion control. Control-A-Video \cite{hu2023videocontrolnetmotionguidedvideotovideotranslation}, for example, builds upon ControlNet's \cite{zhang2023addingconditionalcontroltexttoimage} approach by introducing content priors and leveraging the diffusion model for text-to-video translation. It incorporates both content and motion priors, taking an initial frame as a static content condition and motion information from source videos. While relying on the first frame allows for smoother and more contextually consistent motion, it tackles a different problem than our goal, which is end-to-end T2V generation. 
Other works that perform purely T2V generation, such as Tora~\cite{zhang2024tora} and MotionCtrl~\cite{wang2024motionctrl}, have shown impressive trajectory-controlled generation abilities. 
%
% \noindent 
Tora~\cite{zhang2024tora} employs a Diffusion Transformer and incorporates trajectory conditioning through special modules like the Trajectory Extractor and Motion Guidance Fuser. In contrast, MotionCtrl~\cite{wang2024motionctrl} uses the more traditional U-Net denoising approach and takes as control the camera position and object motion priors. However, these approaches face object-identity issues with objects occasionally displaying shape-shifting or unintended changes in appearance. Our method explicitly models the interactions between objects, which leads to a better perception of object-identity and appearance consistency.

\subsection{Video Generation Dataset}
In recent years both real-world and synthetic video datasets have become essential for advancing video generation. Large-scale real-world datasets such as Panda-70M and WebVid-10M \cite{chen2024panda, Bain21}, provide an extensive collection of video-text pairs which are essential for training large video generative models. Panda-70M specifically captures a wide range of human-centered activities. It also uses automatic captioning methods to enhance the text descriptions. WebVid-10M, on the other hand, consists of videos and captions collected from the internet. Its captions are always considered noisy, which can directly affect generation. 
% \xw{what's online sourced data?}

In addition to real-world datasets, synthetic datasets \cite{greff2022kubric, yi2020clevrercollisioneventsvideo} have recently become critical for controlled video generation. MOVi \cite{greff2022kubric} offers high-quality, 3D-rendered synthetic videos useful for controlled video generation, though its very short video clips limit applications requiring longer temporal patterns. We incorporate the MOVi dataset in our experimentation but we generate longer videos ourselves using their pipeline.
CLEVERER \cite{yi2020clevrercollisioneventsvideo} provides long videos with descriptive texts. While it is an effective simple dataset that can be useful to evaluate complex diffusion models, it lacks certain complex interactions, such as 3D projectile motion that we wish to learn as part of the trajectory-oriented video generation goal.

% \subsection{}
% Related works goes here. However, also
\section{Method}
\label{sec:method}

\subsection{Preliminary}
 We follow the recent works \cite{ma2024latte, zhang2024tora} to encode the video into a latent space during training and inference. Denote the input video as $v_{1:N_0}$ with frame number $N_0$. The video is encoded and decoded through a VAE, as $z_{1:N}=\mathcal{E}(v_{1:N_0}), \hat{v}_{1:N_0}=\mathcal{D}(z_{1:N})$, where $z_{1:N}$ is the video latent with compressed temporal length $N$ and $\hat{v}_{1:N_0}$ is the reconstructed video. During training, we add random noise to the video latent as
 \begin{equation}
     z_{1:N}^{t} = \sqrt{\alpha_t}z_{1:N} + \sqrt{1-\alpha_t}\epsilon
 \end{equation}
where $0=\alpha_{T}<...<\alpha_{t}...<\alpha_{0}=1$ are hyper-parameters for the diffusion scheduler, $z_{1:N}^t$ is the latent at timestamp $t$, and the added noise $\epsilon\sim \mathcal{N}(0,I)$.

During the denoising process, a neural network $\epsilon_{\theta}(z_{1:\check{N}}^t, t, c)$ is applied to predict the added noise, where $c$ is the conditions added to the model. The model training follows the objective:
\begin{equation}
\min_{\theta}E_{z_{1:N}, \epsilon\sim \mathcal{N}(0,I), T\sim U(0,T)}\|\epsilon-\epsilon_{\theta}(z_{1:N}^t, t, c)\|
\end{equation}

% \subsection{Diffusion Transformer}
\noindent \textbf{Diffusion Transformer (DiT)} \cite{peebles2023scalable} applies the diffusion process using the Transformer \cite{vaswani2023attentionneed} model as the denoising neural network. Following recent video generative models \cite{opensora, ma2024latte, pku_yuan_lab_and_tuzhan_ai_etc_2024_10948109}, we employ a DiT as our model backbone to achieve long-term video generation ability. The input video latent $z_{1:N}\in \mathbb{R}^{N \times H\times W \times C}$ is patched into spatial-temporal visual tokens as the input of the transformer-based architecture. The patchify process is implemented by a convolution with both stride and kernel size $k$, and kernel number $L$. As such, the patchified latent is denoted as $p\in \mathbb{R}^{N\times H/k\times W/k\times L}$. Each token $o\in \mathbb{R}^{L}$ is regarded as a feature vector with embedding size $L$, so the total number of spatial-temporal tokens is $(T\times H\times W) / k^2$. These tokens will be input into the DiT.

Unlike the widely used UNet architecture, the DiT comprises of multiple sequential spatial and temporal transformers. The main reason for decoupling the spatial and temporal transformers is to reduce the number of tokens handled by each cross and self-attention. For the spatial transformer, the input latent is reshaped to $p_{spatial}\in \mathbb{R}^{N\times \frac{H\times W}{k^2} \times L}$, where $N$ and $\frac{H\times W}{k^2}$ are regarded as batch size and sequence length, respectively. For the temporal attention, the input latent is reshaped to $p_{temporal}\in \mathbb{R}^{\frac{H\times W}{k^2} \times N \times L}$, so $\frac{H\times W}{k^2}$ is regarded as batch size and $N$ is regarded as sequence length. Therefore, this mechanism allows DiT to focus on spatial and temporal dimensions and reduces the number of computations.

\noindent \textbf{Text-conditioning in DiT} Text conditioning is an important component of DiT where typically, the text-encoded information, processed through a tokenizer and text encoder \cite{raffel2020exploring, radford2021learning}, is integrated into the model via cross-attention within the spatial transformer. The primary reason for using a spatial transformer rather than a temporal transformer lies in the alignment issue between textual and visual information. Even with advanced video captioning models, the extracted textual descriptions are not frame-by-frame (see Appendix A for more information). As a result, applying text-based conditioning within a temporal transformer could lead to temporal inconsistencies. 
%Moreover, we observe that encoding sparse pose and object ID in the spatial transformer can interfere with text-based conditioning, reducing model performance. Therefore, in our model, the spatial transformer incorporates both self-attention and cross-attention for injecting text-based conditioning, while the temporal transformer includes self-attention and cross-attention for injecting sparse pose and object ID.
% \xw{same question here, if this is the standard way, then move it to the preliminary section}
% \textcolor{green}{[Ahmad: I added the DIT part as prelim as it is standard but i think text conditioning in this way is not so a separate section seems appropriate] }\textcolor{blue}{reply: merged into preliminary and delete the description of injecting id and pose into temporal transformer part}

\begin{figure}    
    \centering
    \includegraphics[width=0.8\columnwidth]{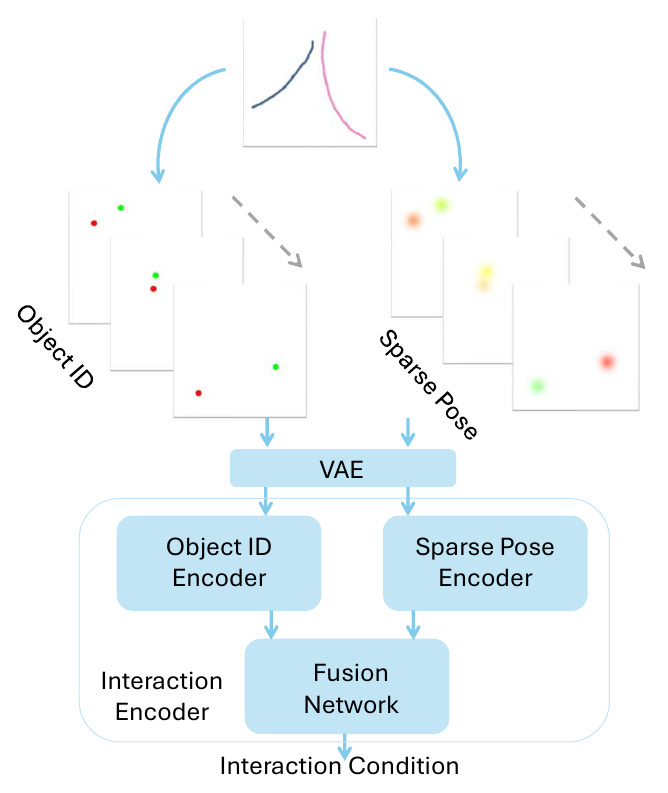}
    \caption{The process of multi-modal interaction encoding. The user-provided input trajectory is first transformed into sparse poses (representing the dynamic information) and object ID maps (containing static and interactive information). Then, two independent encoders are applied to encode their information, and then the fusion network combines the encodings and outputs the interaction condition for the DiT.}
    \label{fig:multi-modal_interaction_encoding}
    \vspace{-5mm}
\end{figure}

\subsection{Object Interaction in Video Generation}
Object interactions in real-world videos are often complex, making video generation a challenging task. We divide object motion information into three major components: dynamic, static, and interactive. Dynamic information refers to the moving objects in the scene, such as a moving car or a running person. Static information encompasses stationary objects within the video. Notably, due to a video's temporal nature, some objects may remain static during certain parts of the video and move in others. Interactive information captures the interactions between objects, such as collisions or bouncing. Additionally, in a two-dimensional video representation, certain forms of `visual interaction,' like crossing or overlapping, can also occur. Previous works fail to account for these complex interaction scenarios, leading to poor performance in depicting realistic object interactions in generated videos.

\subsection{Multi-modal Interaction Encoding}

To capture both dynamic and static information, as well as the interactive information in the video generation model, we propose a multi-modal interaction encoding mechanism, as shown in Fig.~\ref{fig:multi-modal_interaction_encoding}. The input is motion trajectories, and the output is the interaction condition, which will be used to guide the model in generating videos with rich object interaction. %(changing the video object but keeping the trajectory).

%For most videos, the object interaction always contains dynamic information (objects are moving) and stationary information (objects remain stationary or first move and then keep static).

%
%Interactions such as collisions, crossings, and bouncing may occur. 
% Also, there will be interactive information such as collision, crossing, bouncing, etc. 
%
\noindent \textbf{Sparse pose encoding}. Inspired by motion-based diffusion models \cite{wang2024motionctrl, yin2023dragnuwa}, we employ sparse optical flow to encode dynamic interactions in our model. This sparse flow is derived from the input trajectories.
For the trajectory (a list of coordinate points) $(x_i,y_i)_{i:=1:N}$, the difference between adjacent two points is calculated as $(dx_i,dy_j)=(x_i-x_{i-1},y_i-y_{i-1})_{i:=2:N}$, which can also be regarded as the moving speed for the trajectories. Then, following the standard optical flow visualization technique, we use the color wheel to transform the moving direction and speed of the trajectory into the RGB domain. Under this operation, different colors will represent different moving directions and speeds. It is worth noting that static objects lack both direction and speed, meaning they do not exhibit any sparse pose. As a result, sparse pose only represents information about moving objects. Following recent work \cite{wang2024motionctrl}, we apply Gaussian blur to the sparse flow, dispersing it to nearby areas and making it easier for the model to capture.
% In addition, when two objects are very close to each other, the sparse flow will fuse and make it hard to recognize if it is one or two objects, which make it difficult for the model to learn the interaction information.
Furthermore, when two objects are in close proximity, their sparse flows can merge, making it challenging to distinguish between them, To address this, it is necessary to introduce an additional control signal, enabling the model to better understand the interaction.

\noindent \textbf{Object ID encoding}. To encode the stationary and interactive information, we further propose using an additional modal called object ID maps. As shown in Fig. \ref{fig:multi-modal_interaction_encoding}, the object ID maps are also obtained from the input trajectory. However, instead of using color to encode the moving speed, we use it to encode the unique ID for each object. Therefore, even if two trajectories are close to each other, different colors also make them recognizable by the model. We do not consider the object ID overlap because it is very rare in natural video that two objects coincidentally have exactly the same center point. The color of object ID maps keeps constant during the whole video. Therefore, if there are some stationary objects, the model can still recognize their existence and position through ID maps.

\noindent \textbf{Multi-Modal Interaction encoder}. As shown in Fig.\,\ref{fig:multi-modal_interaction_encoding}, the interaction encoder contains two independent encoders to encode sparse poses and object ID maps, respectively, and merge the outputs through a fusion network. The VAE ahead of the interaction encoder is used to encode the interaction information into latent space to align with the visual tokens in the DiT.

\subsection{Video Interaction Dataset}
Evaluating a model's ability to properly generate object interactions is a very challenging task. In real-world datasets, factors such as static video content, video noise, and the complexity of human structures make direct evaluation difficult. Moreover, the domain of real-world data is too broad, and the lack of ground truth further complicates the assessment of generation accuracy. To address these challenges, we propose a new dataset, the Video Interaction (ViN) Dataset, which features diverse scenes with rich object interactions to facilitate more effective evaluation.
%To capture the underlying object interactions, in addition to the visual complexity of real-world videos, our model employs both synthetic and real video data. Unless the data is available as ground truth, videos are captioned with SharedGPT4Video \cite{chen2024sharegpt4video}, their dense optical flow is estimated with Unimatch \cite{xu2023unifying}, and their trajectories are acquired through YoloV7 \cite{wang2023yolov7}. Exact numerical data about the datasets is available in the Table \ref{tab:data}.

\begin{table*}
  \centering
\caption{Information for our proposed ViN dataset.}
  \label{tab:data}
  \begin{tabular}{@{}lccccccc@{}}
    \toprule
    Dataset & \# Videos & Resolution & \# Frames & Caption & Optical flow & Sparse flow & Trajectory \\
    \midrule
    Pool & 20k & 1080x1080 & 120 &Yes (Gen.) & Yes (Gen.) & Yes (Gen.) & Yes (GT) \\
    Dominoes & 5k & 1080x1080 & 152 & Yes (Gen.) & Yes (Gen.) & Yes (Gen.) & Yes (GT) \\
    Football & 5k & 1080x1080 & 230 & Yes (Gen.) & Yes (Gen.) & Yes (Gen.) & Yes (GT) \\
    MOVi-Extended & 20k & 256x256 & 65 & Yes (Gen.) & Yes (GT) & Yes (GT) & Yes (GT) \\
    \bottomrule
    \label{dataset_info}
  \end{tabular}
  \vspace{-8mm}
\end{table*}

The main purpose of the ViN dataset is to offer a high degree of control over the generated outputs and provide high-quality information about objects' locations in each frame. This simplifies the evaluation of generative models, particularly in scenarios involving complex object interactions. The ViN dataset comprises four subsets: the \textbf{Pool dataset}, \textbf{Domino dataset}, \textbf{Football dataset}, and \textbf{MOVi-Extended dataset}. The MOVi-Extended dataset is generated using the Kubric \cite{greff2022kubric} pipeline. Additionally, we generate longer videos within this subset to better support the requirements of video generation tasks.
To increase the dataset diversity, we use Blender \cite{blender} for generating the Pool and MOVi-Extended datasets, while Unity \cite{unity} is employed for generating the Football and Domino datasets. Each engine offers distinct advantages: Blender provides high-quality visual fidelity, while Unity excels at rendering fully animated physics sequences more efficiently. Blender utilizes the Bullet physics engine, whereas Unity relies on the PhysX engine. By simulating physics interactions across datasets using both programs, we aim to diversify the training data and mitigate biases associated with their respective physics implementations. Our approach focuses on creating and utilizing a variety of domain-specific datasets for object interactions, which can be combined with realistic counterparts to enhance model performance and generalization capabilities. An overview of the four generated datasets is given in Table \ref{dataset_info}. When groundtruth data is unavailable, videos are captioned with SharedGPT4Video \cite{chen2024sharegpt4video}, dense optical flow is estimated with Unimatch \cite{xu2023unifying} (see Implementation details for more information), and trajectories are extracted using YoloV7 \cite{wang2023yolov7}.

% Videos are generated with Blender for the Pool and MOVi Extended datasets and Unity for the Football and Dominoes datasets. Blender focuses at producing 3D output with high-detailed visual quality while Unity excels at rendering significantly faster. Through using multiple engines to grasp the physical interactions of the scenes we hope to diversify our training data as well as overcome any bias in the computational methods employed to do the interaction calculations in the background.

To capture longer and more diverse interactions while maintaining simplicity, we generated 65-frame videos using the Kubric \cite{greff2022kubric} MOVi generation pipeline with 1 to 10 objects.  The dataset includes scenarios where objects leave the scene at different times due to factors such as direct impacts or natural movement. Additionally, it features objects transitioning from movement to a stationary state and receding into the background.

% This has enabled the observation of objects which leave the scene at different times as well as for different reasons like direct impact or natural movement. At the same time we can observe object shrinking in the background and changing of objects' state from moving to stationary.

The Pool dataset excels in visual quality and supports complex interactions involving up to 16 objects. It features objects that change direction multiple times due to collisions with the edges of the table or other pool balls. The interactions also include the disappearance of objects as they fall into pockets.

The Football dataset focuses on learning visually complex elements, featuring a goalkeeper and a player interacting with a ball. Its structured and realistic backgrounds enhance the resemblance to real-world settings.

To scale up the number of objects to be tracked, we introduce the Domino dataset. This dataset presents simple interactions and scene settings while offering a challenging scenario involving up to 20 objects to be accounted for.

To preserve visual realism in real-world settings, we utilize a subset of Panda-70M \cite{chen2024panda}, complemented by generated sparse flow, optical flow, and trajectory estimates. To balance the quality of the data and minimize the gap between real and synthetic datasets, this real subset is similar in size to the generated datasets.

% the generated datasets, this dataset has a similar size

% This dataset is similar in size with the synthetic datasets in order to balance the quality and quantity of our data.  

% \subsection{ID-based DiT Model}

% The overall architecture is shown in Fig. \ref{fig:overall_structure}. We follow the model backbone from Open-Sora-Plan \cite{pku_yuan_lab_and_tuzhan_ai_etc_2024_10948109} and Latte \cite{ma2024latte} model. Due to 

% \subsubsection{Multi-Modal Encoder for trajectory control}
% Although some recent works \cite{zhang2024tora, wang2024motionctrl, yin2023dragnuwa} emphasis the importance 
% As shown in Fig.\ref{fig:overall_structure}, 

% \subsection{ID Control Injection Mechanism}
% \begin{figure*}[h]
%     \centering
%     \includegraphics[width=0.5\textwidth]{./Images/overall_disgram.pdf}
%     \caption{This is an example caption for the image.}
%     \label{fig:overall_structure}
% \end{figure*}

\subsection{Trajectory Evaluation}
To quantitatively evaluate the accuracy of generated videos in terms of how closely the output follows the input trajectory, we propose a matching trajectory evaluation metric (MTEM). This metric compares the ground truth trajectory, provided as input, with the trajectories extracted from the generated output videos, and yields a similarity score between the two sets of trajectories.

To extract a set of trajectories from the generated output we use an object detection model on each frame which provides us the coordinates of detected objects along with their corresponding IDs, allowing us to track which object appears in each frame. As a result, we can construct the moving trajectories of each object. However, due to the imperfect quality of generated videos, objects may occasionally disappear and reappear later on. For reasons that will become clear later, we require trajectories to be defined across all frames, even those where an object is temporarily missing. To address this, if an object disappears but has been detected previously, we assign its coordinates in the missing frames to those from its most recent appearance.

With the extracted trajectories from the generated videos we now evaluate this trajectory using our MTEM metric; Formally, let $T_1$ and $T_2$ be two sets of trajectories, where each trajectory is a sequence of points $\{(x_i, y_i) : i \in [0, F]\}$, and $F$ denotes the total number of frames in the video. Ideally, we would have $|T_1| = |T_2|$, and each trajectory in $T_1$ would directly correspond to a trajectory in $T_2$. However, this is not always the case, and even when $|T_1| = |T_2|$, it may not be clear which element in $T_1$ best corresponds to an element in $T_2$.

To address this, we model $T_1 \cup T_2$ as a complete bipartite graph, where the nodes of the first partition correspond to the trajectories in $T_1$ and the nodes in the second partition correspond to the trajectories in $T_2$. The weight of the edge between any two nodes $t_1 \in T_1$ and $t_2 \in T_2$ is given by the distance $d(t_1, t_2)$, defined as:

\[
d(t_1, t_2) = \sum_{i = 1}^{F} \left\| \left(x_i^{(1)}, y_i^{(1)}\right) - \left(x_i^{(2)}, y_i^{(2)}\right) \right\|_2^2,
\]

where $\left(x_i^{(1)}, y_i^{(1)}\right)$ and $\left(x_i^{(2)}, y_i^{(2)}\right)$ represent the coordinates of the $i$-th frame in trajectories $t_1$ and $t_2$, respectively.

Next, we apply the Hungarian matching algorithm to find the minimum-weight matching between the trajectories in $T_1$ and $T_2$. The output of the Hungarian algorithm is a set of matched trajectory pairs with the minimum total distance, providing a quantitative measure of the similarity between the generated and ground truth trajectories.

\begin{table*}[h]
  \centering
  \small
  \caption{User study for controlled movement (C) and interaction realism (R). The reported numbers, excluding the MOVi-Extended are the result of 1000 opinions split equally between the categories and models.}
  \begin{tabular}{@{} c c c c c c c @{}}
    \toprule
    \\ Model & \multicolumn{2}{c}{MOVi-Extended} & \multicolumn{2}{c}{Domino} & \multicolumn{2}{c}{Pool} \\
    \cmidrule(lr){2-3} \cmidrule(lr){4-5} \cmidrule(lr){6-7}
    & R & C & R & C & R & C  \\
    \midrule
    InTraGen & 3.24 \(\pm\) 1.41 & 4.48 \(\pm\) 0.85 & 4.40 \(\pm\) 0.68 & 4.63 \(\pm\) 0.54 & 3.77 \(\pm\) 1.16 & 4.20 \(\pm\) 1.22 \\ 
    Original videos & 4.91 \(\pm\) 0.35 & 4.53 \(\pm\) 0.83 & 4.57 \(\pm\) 0.65 & 4.66 \(\pm\) 0.49 & 4.40 \(\pm\) 1.00 & 4.43 \(\pm\) 1.09 \\
    \bottomrule
  \end{tabular}
  
  \label{tab:useralldata}
\end{table*}

\begin{table*}
  \centering
  \small
  \caption{Quantitative comparison between Open-Sora-Plan \cite{pku_yuan_lab_and_tuzhan_ai_etc_2024_10948109}, our model and our model without object id.}
  \begin{tabular}{@{}lccccc@{}}
    \toprule
    Model & MTEM(\%)\(\downarrow\) & SSIM(\%)\(\uparrow\) & PSNR\(\uparrow\) & LPIPS(\%)\(\downarrow\) & FID\(\downarrow\) \\
    \midrule
    Open-Sora-Plan\cite{pku_yuan_lab_and_tuzhan_ai_etc_2024_10948109} & 22.17 & 95.42 \(\pm\) 1.70 & 24.42 \(\pm\) 2.42 & 21.39 \(\pm\) 4.57 & 40.65 \\
    Sparse pose & 10.38 & 96.35 \(\pm\) 1.22 & 25.95 \(\pm\) 2.15 & 14.21 \(\pm\) 3.05 & 36.65 \\
    InTraGen (Sparse pose + Object id) & \textbf{9.40} & \textbf{96.51 \(\pm\) 1.19} & \textbf{26.35 \(\pm\) 2.31} & \textbf{12.61 \(\pm\) 2.48} & \textbf{26.37} \\
    \bottomrule
  \end{tabular}
  
  \label{tab:metrics}
  \vspace{-5mm}
\end{table*}

\begin{table}
  \centering
  \small
  \caption{User study for interaction realism on the MOVi-Extended dataset. Results are based on 600 opinions split equally between the four models.}
  \vspace{-2mm}
  \begin{tabular}{@{}cc@{}}
    \toprule
    Model & MOVi-Extended realism \\
    \midrule
    Open-Sora-Plan &  0.66 \(\pm\) 1.16 \\
    Sparse pose & 2.01 \(\pm\) 1.76 \\
    InTraGen (Sparse pose + Object id) & \textbf{3.24 \(\pm\) 1.41} \\
    \bottomrule
    Original videos & 4.91 \(\pm\) 0.35 \\
    \bottomrule
  \end{tabular}
  \label{tab:movirealism}
  \vspace{-5mm}
\end{table}

\section{Experiments}

\subsection{Implementation Details}
Our model is built on top of the open-source Open-Sora-Plan repository \cite{pku_yuan_lab_and_tuzhan_ai_etc_2024_10948109}, which is primarily based on the model architecture of Latte \cite{ma2024latte}. To effectively capture rich motion information, we follow the dense optical flow pre-training strategy proposed by MotionCtrl~\cite{wang2024motionctrl}. This approach involves initially training our model on a dataset paired with dense optical flow data. We use Unimatch \cite{xu2023unifying} to extract the dense optical flow and conduct our experiments on six different datasets. First, we evaluate the performance of our model on the four subsets of the proposed ViN dataset to demonstrate its enhanced ability to generate realistic object interactions. Moreover, we evaluate our model on two real-world datasets, including the Panda-70M \cite{chen2024panda} and SoccerNet \cite{deliege2021soccernet}. For Panda-70M, we cleaned 50K videos, and for SoccerNet, we extracted 18K video clips.

\noindent \textbf{Metrics:} we used the proposed matching trajectory evaluation metric (MTEM) to evaluate the interaction effect. Additionally, we employed widely adopted metrics for generative model, including Peak Signal-to-Noise Ratio (PSNR), Structural Similarity Index Measure (SSIM), Learned Perceptual Image Patch Similarity (LPIPS), and Fréchet Image Distance (FID) to further evaluate our model performance.

\noindent \textbf{Model training:} We used 8 NVIDIA A100 GPUs with 40GB memory for training. For the quantitative comparison, each model was trained for 12,000 steps using the Adam optimizer with a learning rate of $2\times 10^{-5}$. The pre-training stage on dense optical flow also consisted of 12,000 steps with the same optimization settings. Additional implementation details can be found in Appendix C.

\subsection{Quantitative comparison}
\label{sec_quantitative}
We compute the following metrics (as shown in Table \ref{tab:metrics}): MTEM, SSIM, PSNR, LPIPS, and FID, to compare Open-Sora-Plan \cite{pku_yuan_lab_and_tuzhan_ai_etc_2024_10948109} with our model. To evaluate the impact of encoding static and interactive information, we also test a variant of our model that removes the object ID component, retaining only the sparse pose. This approach is consistent with other trajectory-control models \cite{wang2024motionctrl, yin2023dragnuwa}, which also use sparse pose to capture motion information. Our results show that incorporating object ID maps during training improves all video evaluation metrics.

\section{Qualitive Comparison}
\begin{figure}    
    \includegraphics[width=\columnwidth]{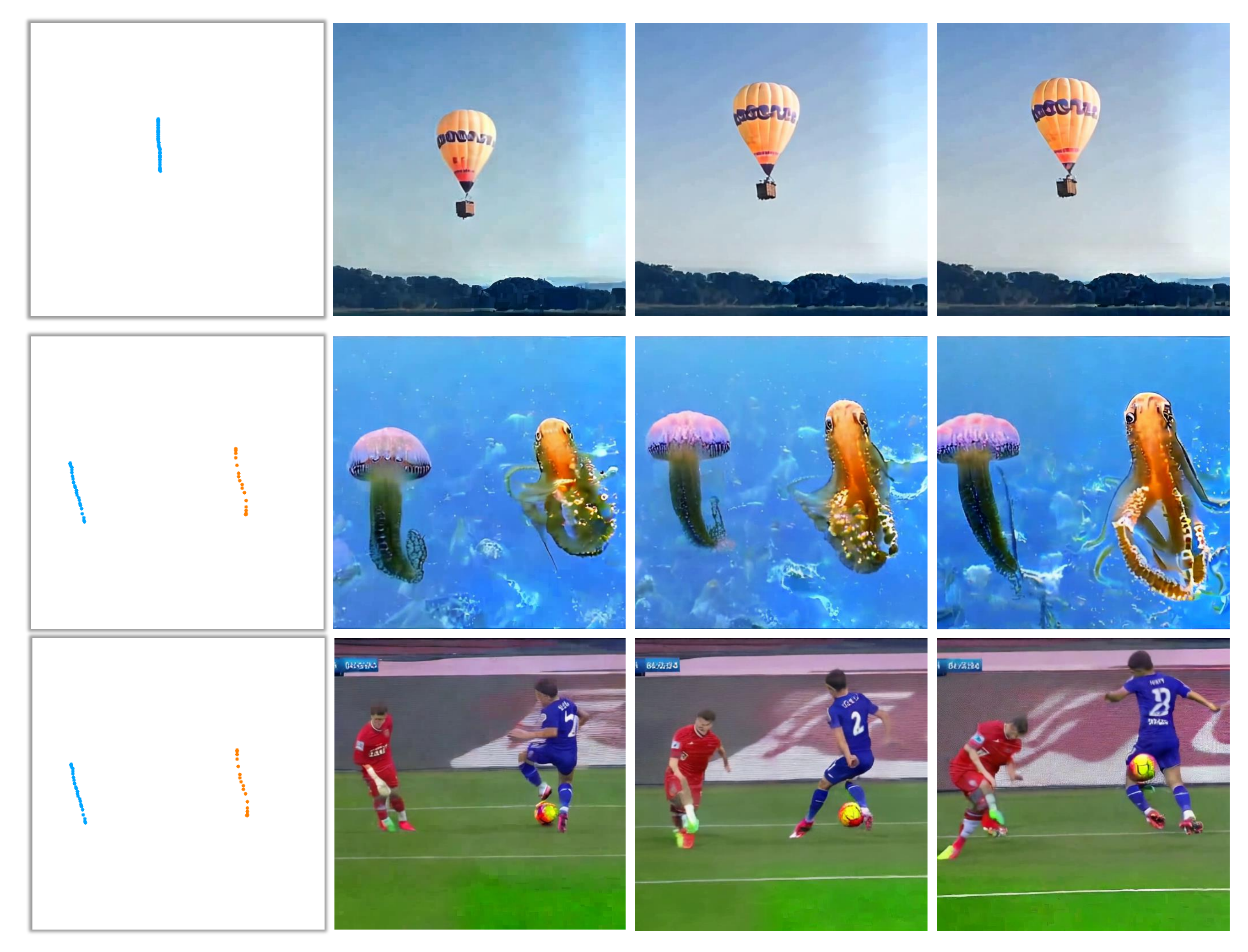}
    \caption{Qualitive evaluation of our model on Panda-70M and SoccerNet dataset}
    \label{fig:real_dataset}
    \vspace{-5mm}
\end{figure}

\begin{figure}    
    \includegraphics[width=0.9\columnwidth]{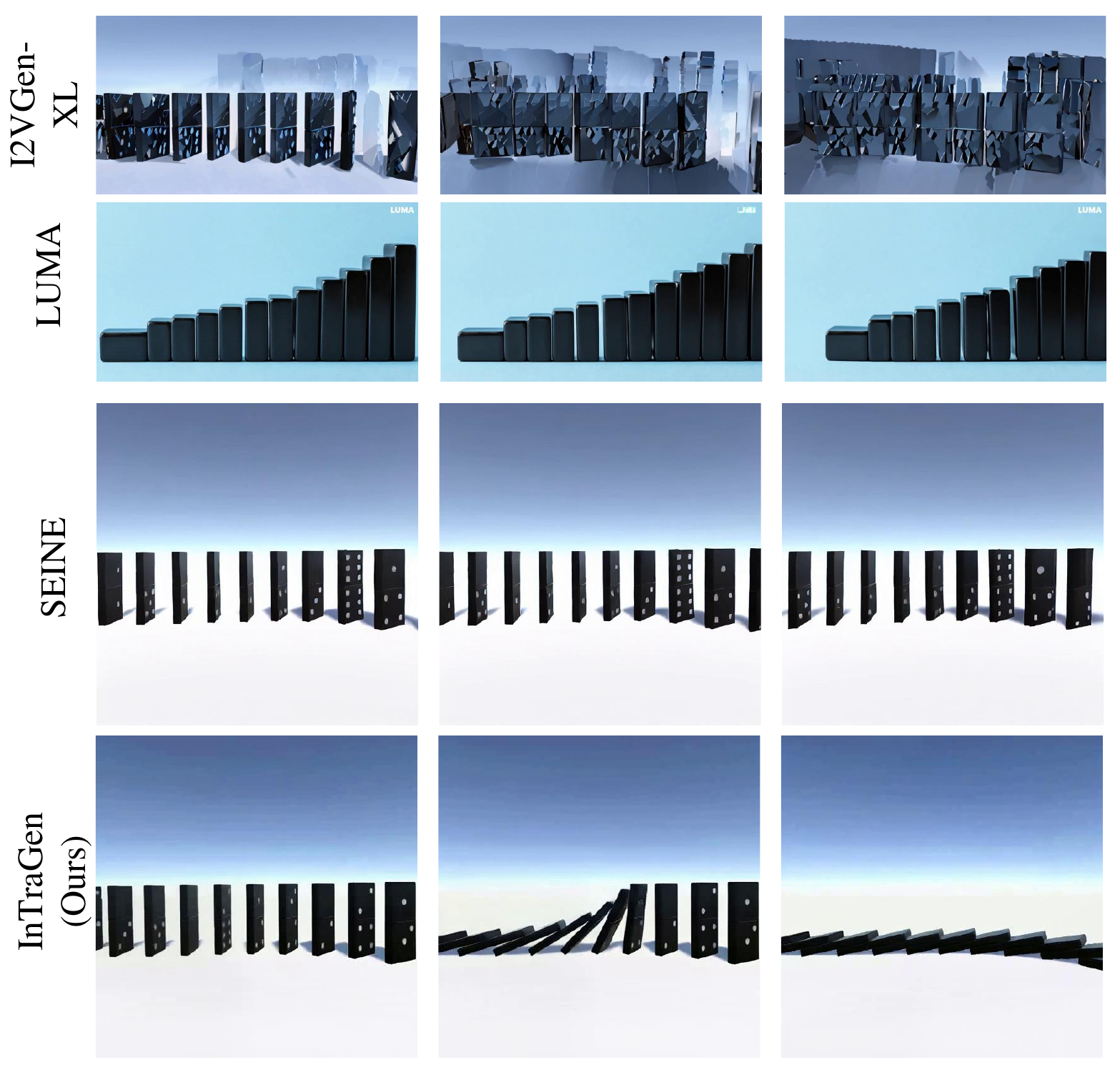}
    \vspace{-5mm}
    \caption{Comparison with different advanced and commercial models on an example from our dominos dataset. All other methods including I2VGen-XL\cite{zhang2023i2vgen}, LUMA \cite{luma}, and SEINE \cite{chen2023seine} are failing in this scene. Our method shows the most realistic results in this interactive scenario. For SIENE and I2VGen-XL, the first frame from the ground truth is additionally passed.}
    \label{fig:compare_with_commercial}
    \vspace{-5mm}
\end{figure}

The qualitative results of the proposed InTraGen can be seen in Fig.\,\ref{first_page_fig}, Fig.\,\ref{fig:real_dataset} and Fig.\,\ref{fig:compare_with_commercial}. In Fig.\,\ref{first_page_fig}, we show the results on the MoVi-Extended, Pool Game, and Domino datasets. Its performance shows that the model can generate excellent interactive performance, which is aligned with our quantitative results in Section \ref{sec_quantitative}. In Fig.\,\ref{fig:real_dataset}, the control and interaction results in the real dataset can also be observed, with aesthetically pleasing videos. More video results can be found in our supplementary materials. In Fig.\,\ref{fig:compare_with_commercial}, we show the comparison with different advanced and commercial models for the dominos dataset. All other methods including I2VGen-XL\cite{zhang2023i2vgen}, LUMA \cite{luma}, and SEINE \cite{chen2023seine} are failing in this scene. Our method shows the most realistic results in this interactive scenario.

\subsection{Ablation Study}
In order to validate the significance of different components,  we ablated our model whenever possible. In particular, we compared our final method against the baseline, both with and without the object ID. Note that ablation without sparse pose is not necessary since our object ID also includes pose information indirectly. The obtained results can be found in Table~\ref{tab:metrics} and the supplementary materials.

\subsection{User study}

We conducted two user studies - on multi-object interaction realism and adherence to a predefined trajectory. Summary of he results can be found in Tables \ref{tab:useralldata} and \ref{tab:movirealism}. In both studies, the videos belong to one of three categories: MOVi-style videos, domino videos, and pool videos. Participants were asked to evaluate the videos on a scale from 0 (no realism) to 5 (highly realistic).

In the first study, we measured the multi-object interaction realism of MOVi-style videos, which are extracted from the following sources: 1) The original dataset; 2) Open-Sora-Plan; 3) Our sparse pose conditioned model; 4) Our sparse pose and object id conditioned model. In this setting, we have 600 opinions split equally between the four models. Results indicate an improvement when sparse pose conditioning is present, which further increases when object ID to distinguish each object is introduced.

In the second study, we evaluated both interaction realism (realism) and adherence to a predefined trajectory (control) from the MOVi-style videos, Domino dataset, and Pool dataset. The comparisons shown in Table \ref{tab:useralldata} include cumulatively 1000 opinions split equally for the categories in table, excluding the MOVi-Extended realism score which is computed through the first study. Results indicate strong ability to simulate diverse domain-specific environments by introducing additional conditioning signals to the model.

\section{Conclusion}
In this work, we addressed the key challenge of text-to-video generation by introducing ``InTraGen", a novel trajectory-controlled video generation framework designed to enhance interaction quality between objects. By leveraging trajectories as a flexible and precise control signal, our method captures dynamic, static, and interactive information with greater accuracy compared to existing approaches. 
To train and evaluate our framework, we created four diverse datasets with rich object interaction scenarios, accompanied by a novel interaction quality metric, enabling a more robust assessment of video generation models. Our results demonstrate significant improvements in both visual fidelity and interaction quality, showcasing the effectiveness of our approach in tackling the complexities of temporal reasoning and object interaction.

\noindent \textbf{Acknowledgements.} This research was partially funded by the Ministry of Education and Science of Bulgaria (support for INSAIT, part of the Bulgarian National Roadmap for Research Infrastructure).

\newpage
{
    \small
    \bibliographystyle{ieeenat_fullname}
    \bibliography{main}
}

\newpage

\appendix
\noindent{\Large \textbf{{Appendix}}}

\section{Limitation of Video Caption Model}
\begin{figure*}[h]
    \centering
    \includegraphics[width=\linewidth]{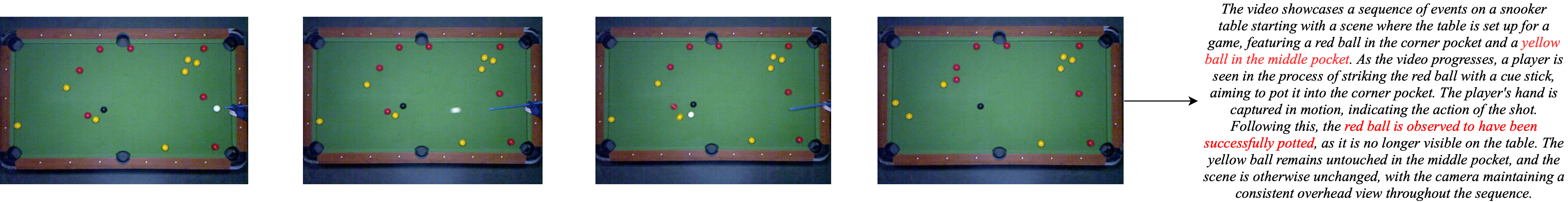}
    \caption{the ShareGPT4Video \cite{chen2024sharegpt4video} creates long captions with irrelevant and confusing information for the simple pool game video.}
    \label{caption_problem}
\end{figure*}

Current video captioning models are unable to capture complex scenes and detailed temporal relationships in the videos. These models often struggle with accurately representing nuanced interactions among multiple objects, which can lead to misleading or incomplete descriptions of the video's content. As shown in Figure \ref{caption_problem}, the ShareGPT4Video \cite{chen2024sharegpt4video} creates long captions with irrelevant and confusing information for the simple pool game video.

\section{Limitation about Text Encoder in Video Generation}

% 1. problem of the caption
% 2. problem of the text encoder
% 3. more experiments
%  a. rendered soccer
%  b. soccer net
%  c. part traj
%  b. view point
%  e. more visualization

% 4. falling cases

% 5. more implementation

\begin{figure}[h]  
    \centering
    \includegraphics[width=0.8\columnwidth]{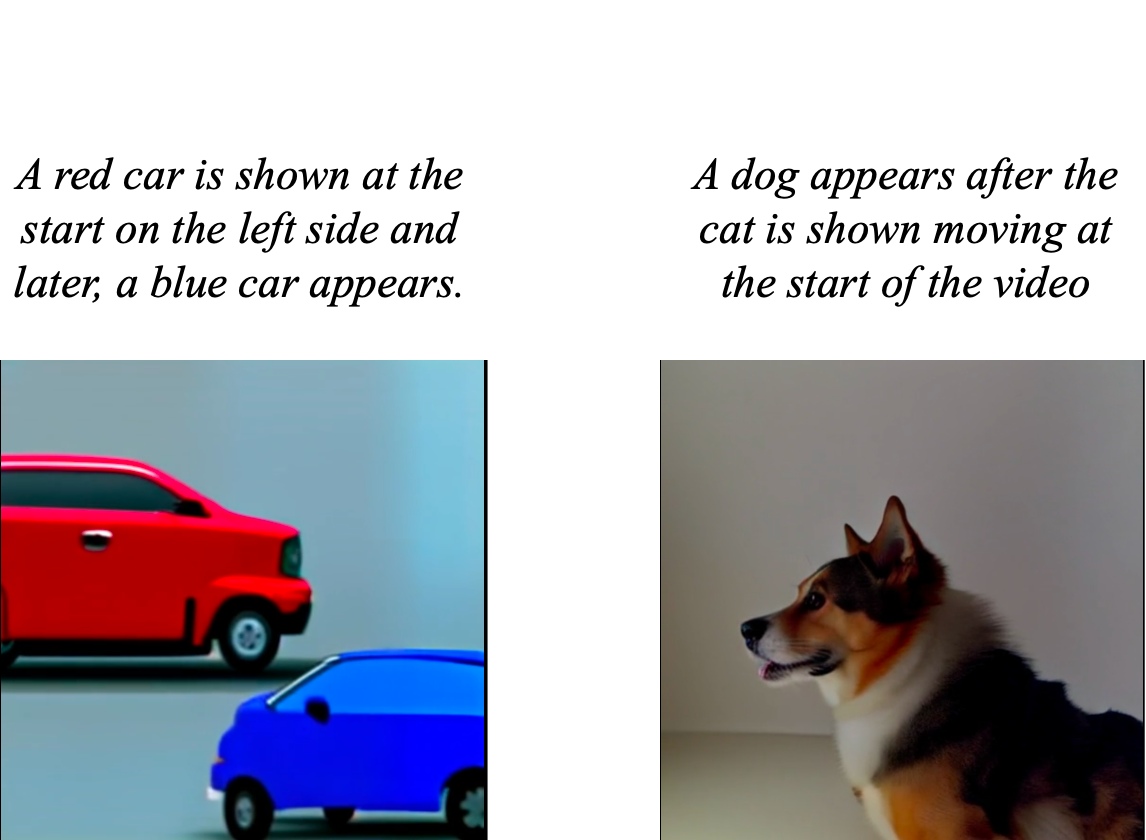}
    \caption{Consequence of a poor text encoder: Lack of temporal understanding of the caption resulting in wrong first frames.}
    \label{text_enc_problem}
    
\end{figure}

Advanced text encoders are struggling to guide the video generation model even with fine-grained text descriptions. This significantly impacts the overall performance of the video generation task. One key limitation is the inability of the text encoder to properly encode temporal information in the text, which results in incoherent video outputs that fail to capture the intended dynamics of the scenes. We show such a failure case with multiple examples. We prepare multiple text prompts, with each prompt describing a simple temporal event such as an object is present at the start frame and another object joins in later frames. However, the generated video always shows both objects in the first frame. We show the caption and first frame of each of these examples in Figure \ref{text_enc_problem}.

\begin{figure}[h]
    \centering
    \includegraphics[width=\columnwidth]{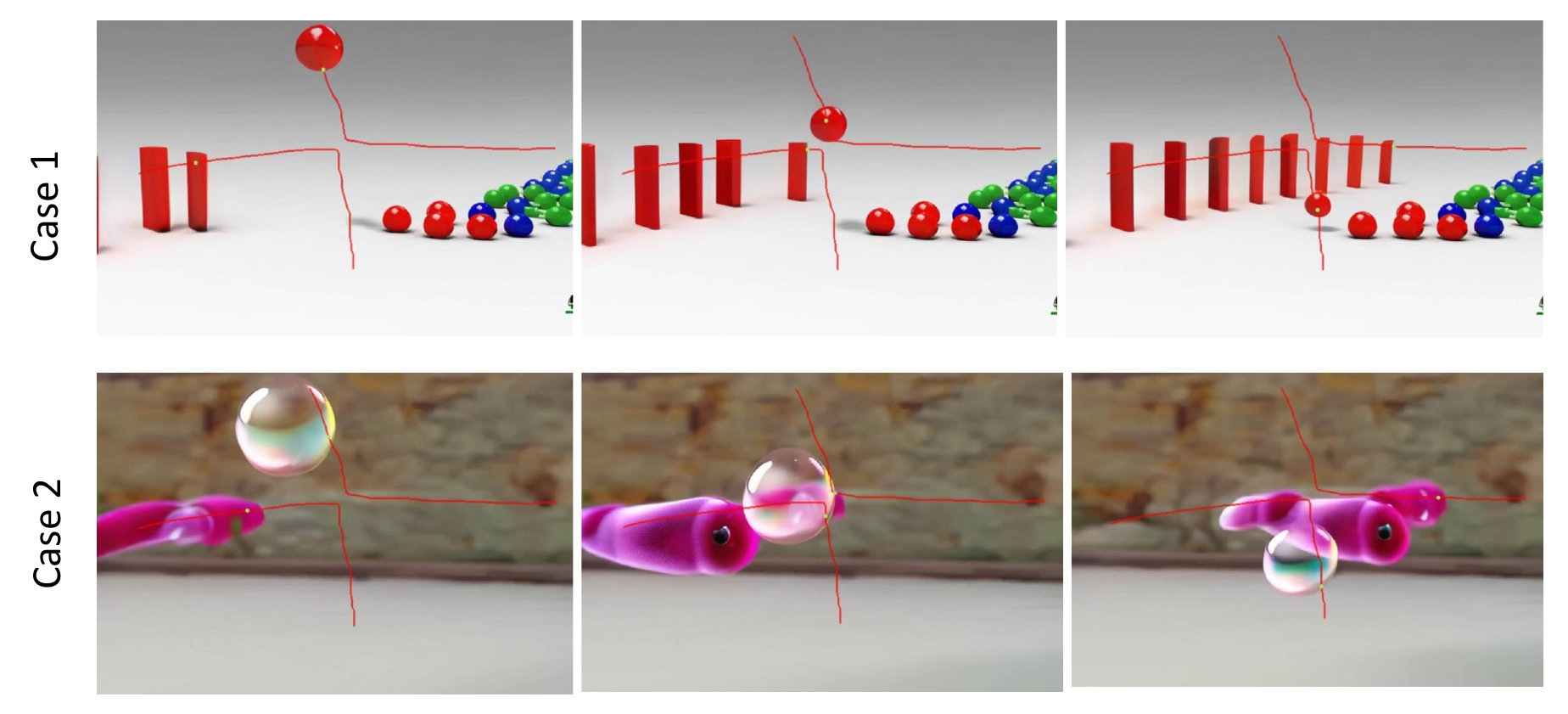}
    \caption{Tora \cite{zhang2024tora} generates videos with poor object interaction. From the cases, we can find that objects mistakenly exchange trajectories when they are close to each other.}
    \label{tora_problem}
\end{figure}

\section{More Experiments}

\subsection{More Visualization on Domino Dataset}
\begin{figure*}[h]
    \centering
    \includegraphics[width=\linewidth]{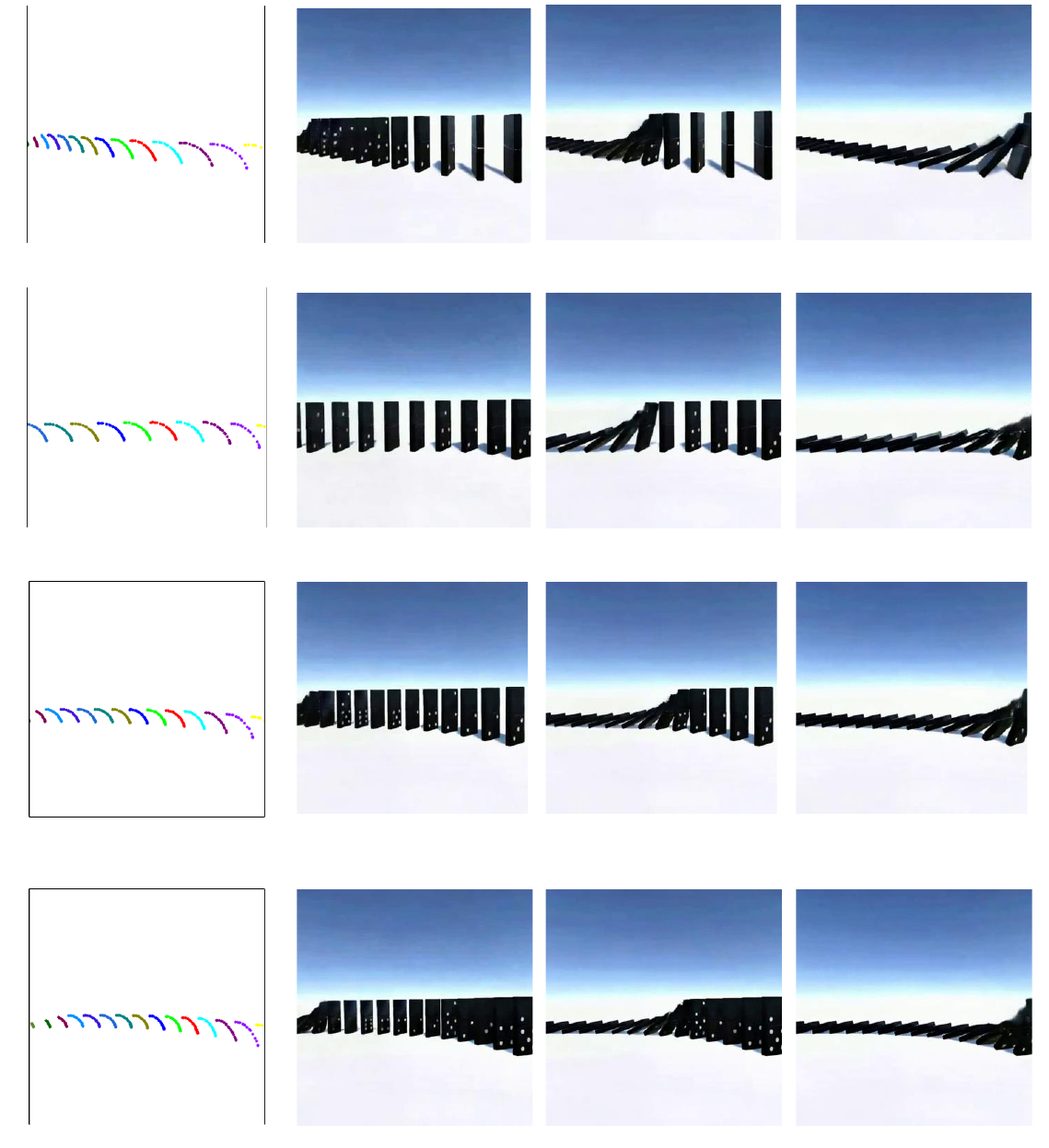}
    \caption{Generated video visualization of InTraGen trained on proposed domino dataset}
    \label{domino_more}
\end{figure*}

We provide more visualization for our proposed domino dataset in Fig.\,\ref{domino_more}. For the dataset, the trajectory control from different viewpoints (angle of the camera to capture the domino falling) and distances (distance between the camera to the domino) are provided. From the figure, InTraGen can always generate results with good performance.

\subsection{More Visualization on Pool Game Dataset}
\begin{figure*}[h]
    \centering
    \includegraphics[width=\linewidth]{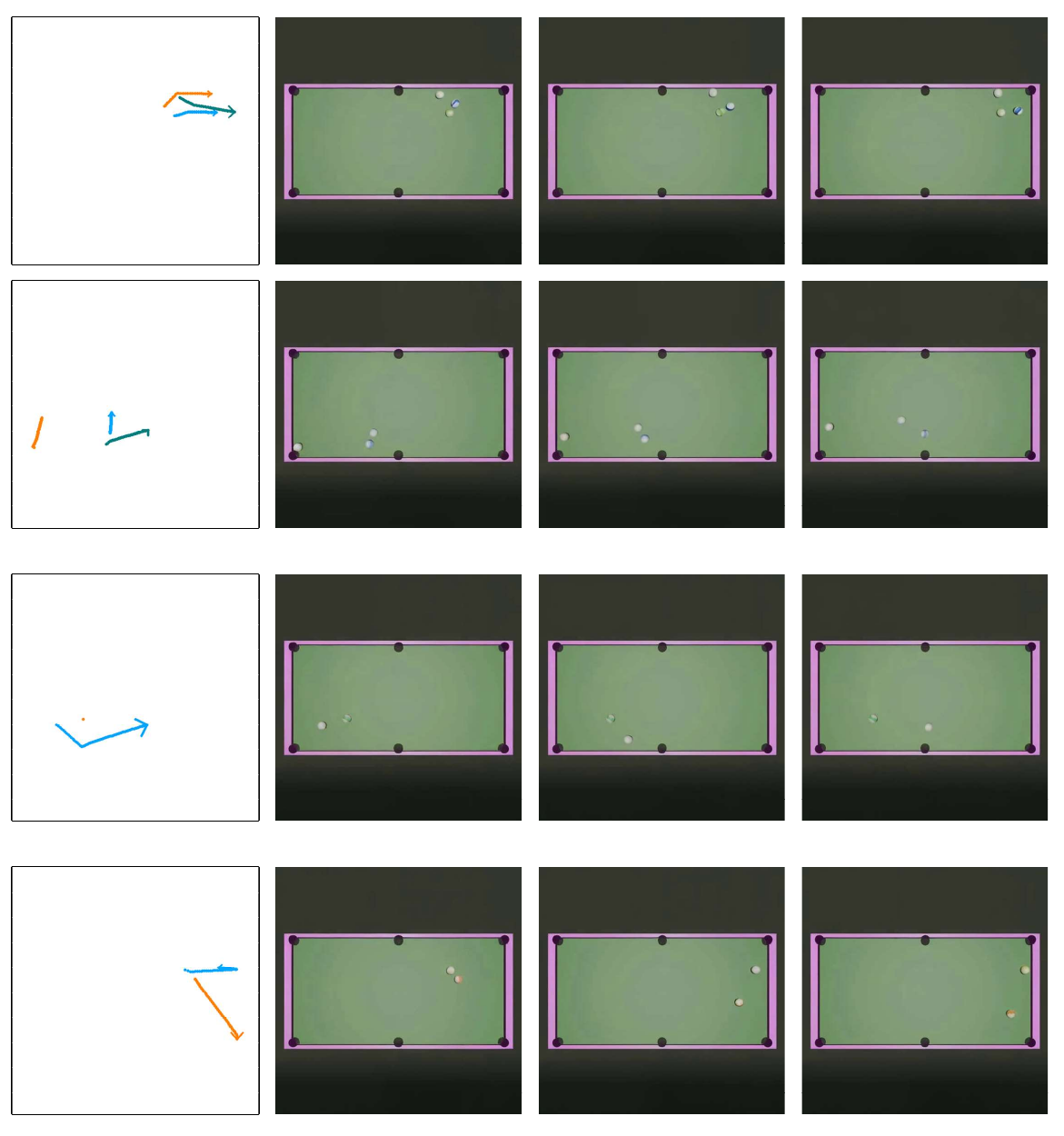}
    \caption{Generated video visualization of InTraGen trained on proposed pool game dataset}
    \label{more_bigballs}
\end{figure*}
More visualization of the pool game dataset can be found in Fig.\,\ref{more_bigballs}. We provide the trajectories of multi-billard interaction. The results also show good performance in terms of temporal consistency and interaction realism.

\subsection{More Visualization on Football Dataset}
\begin{figure*}[h]
    \centering
    \includegraphics[width=\linewidth]{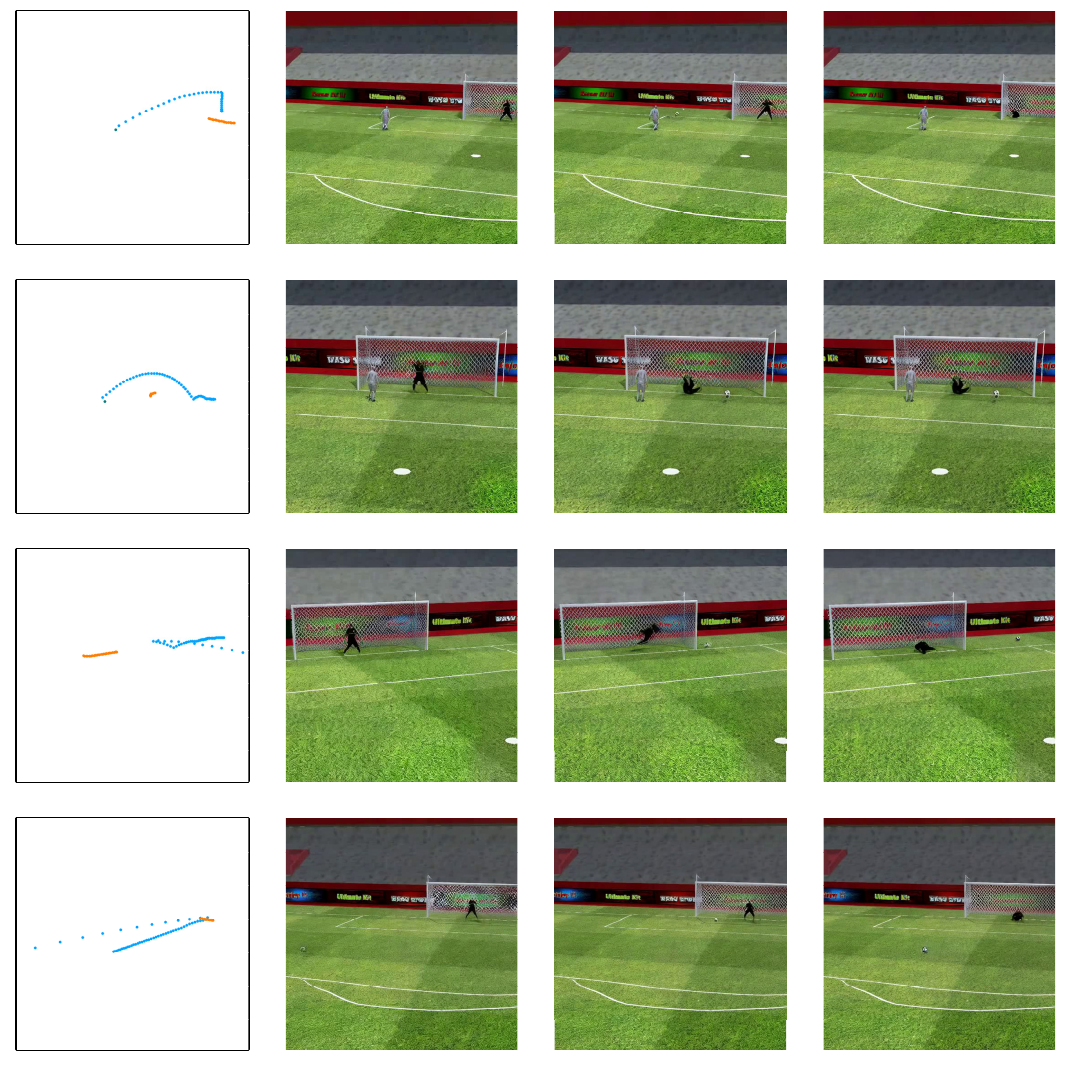}
    \caption{Generated video visualization of InTraGen trained on proposed football dataset}
    \label{more_football}
\end{figure*}
More results for the proposed football dataset are shown in Fig.\,\ref{more_football}. The dataset contains the player, ball, and goalkeeper (they do not necessarily appear in the video at the same time). In the dataset, the ball sometimes flies very fast and its size is very small. However, the proposed InTraGen can still generate good videos with accurate trajectories for the flying ball.

\subsection{More Visualization on Extended-MoVi Dataset}
\begin{figure*}[h]
    \centering
    \includegraphics[width=\linewidth]{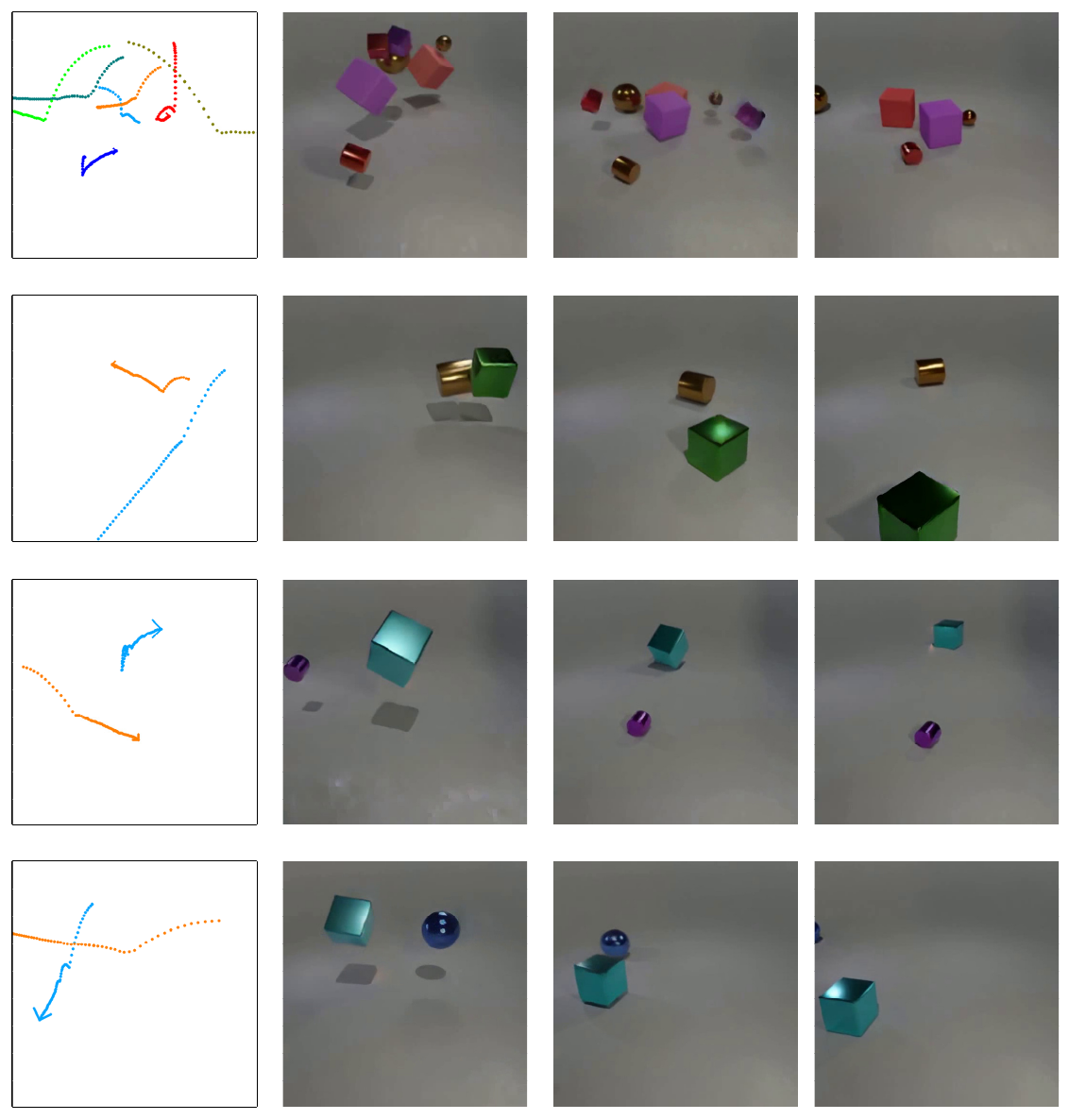}
    \caption{Generated video visualization of InTraGen trained on proposed Extended-MoVi dataset}
    \label{more_cubric}
\end{figure*}
We also provide more visualization for the MoVi-Extended dataset in Fig.\,\ref{more_cubric}. In the first column, the InTraGen can still achieve excellent performance even with many trajectories. For other cases with crossing or collision, the performance is still excellent.

\subsection{More Visualization on SoccerNet Dataset}
\begin{figure*}[h]
    \centering
    \includegraphics[width=\linewidth]{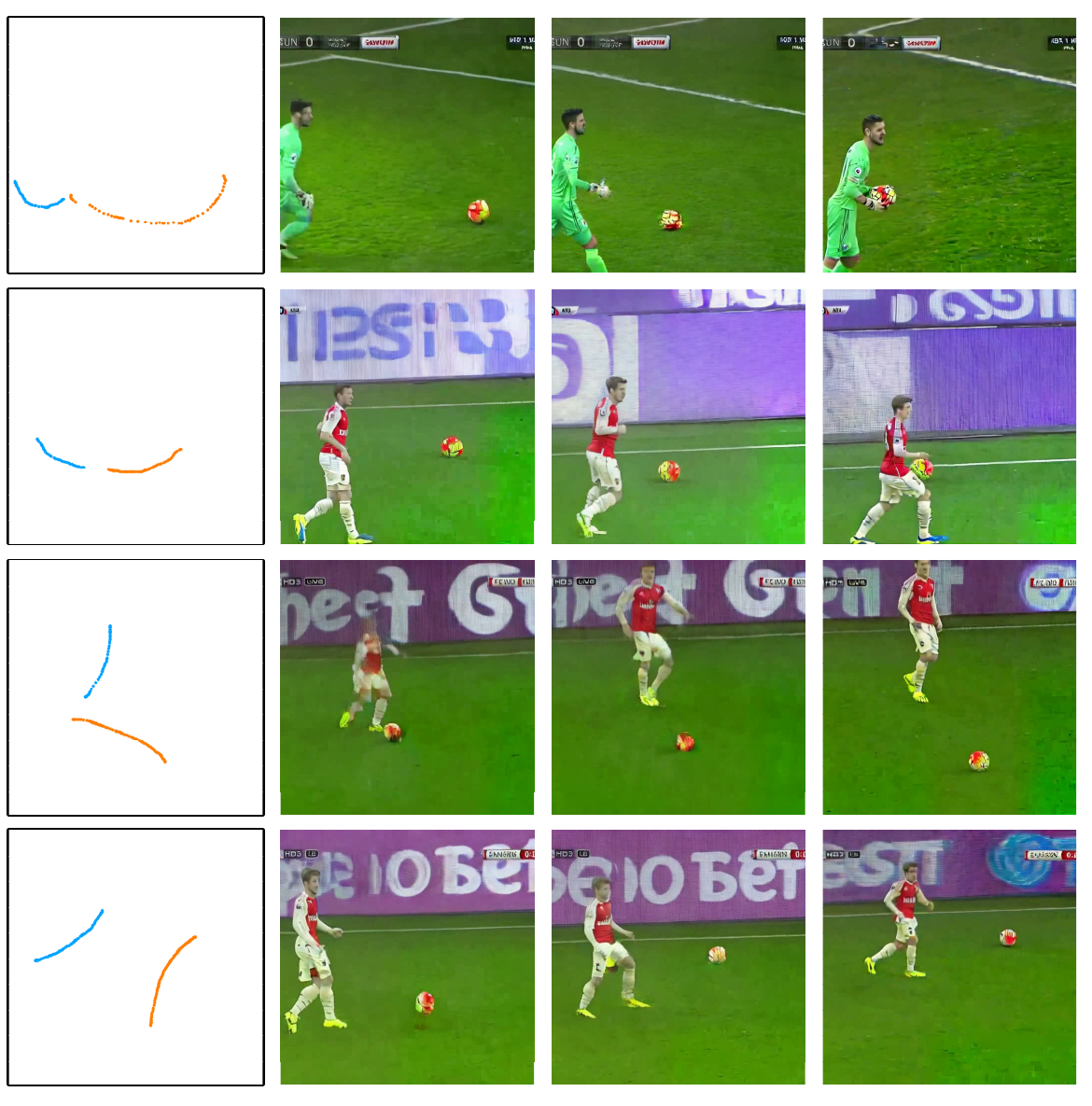}
    \caption{Generated video visualization of InTraGen trained on SoccerNet dataset}
    \label{more_SoccerNet}
\end{figure*}
More performance for InTraGen on SoccerNet dataset \cite{deliege2021soccernet} is shown in Fig.\,\ref{more_SoccerNet}. This is quite a challenging dataset because of complex human motion, background audience, and video noise. However, given different trajectories, the model can well understand their meaning and generate reasonable interaction videos, with fine human motion and ball trajectories.

\subsection{Multi-view Generation}

\begin{figure*}[h]
    \centering
    \includegraphics[width=\linewidth]{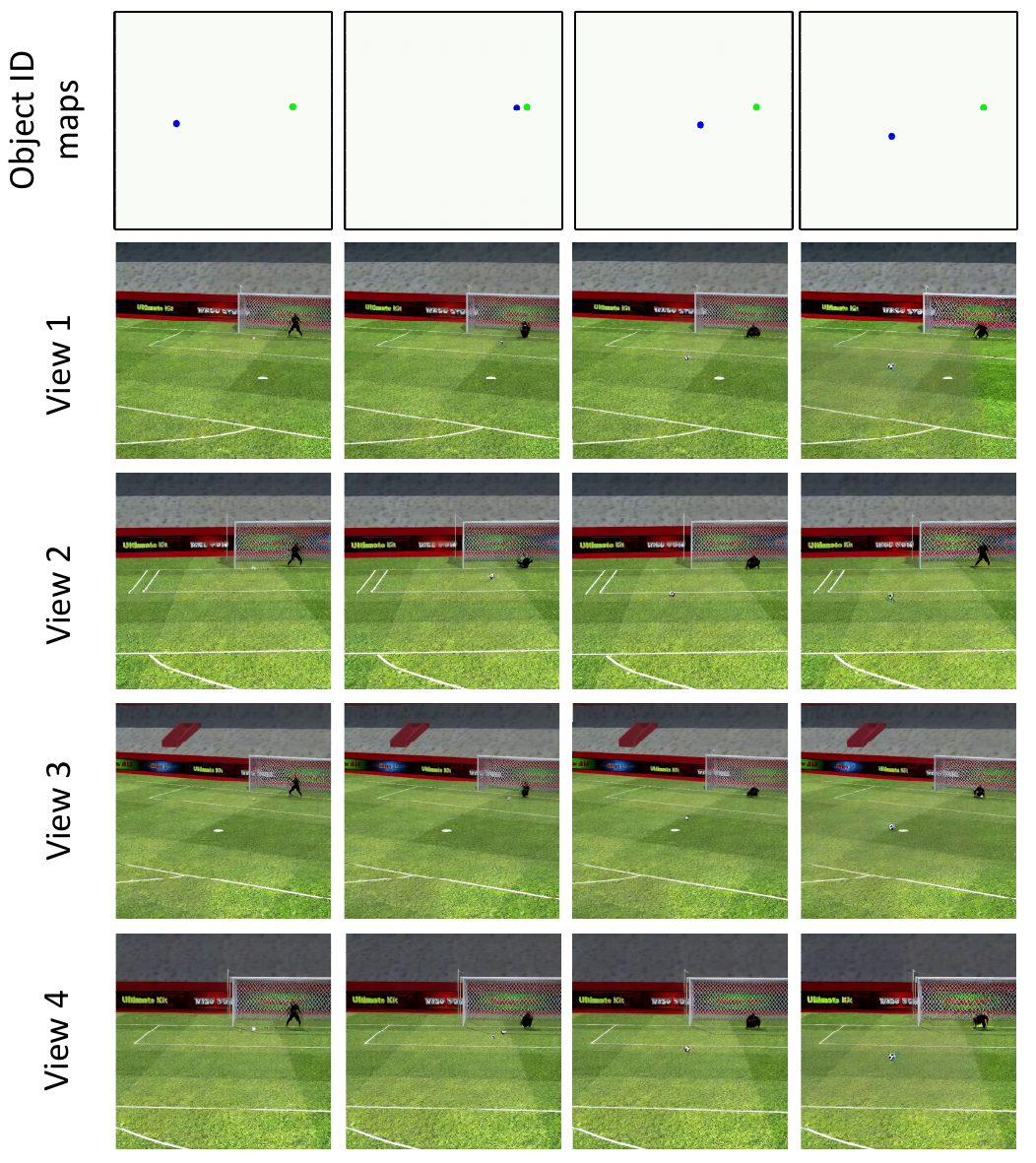}
    \caption{Multi-view generation results of InTraGen.}
    \label{fig:enter-label}
\end{figure*}

The InTraGen can well fit with the multi-view dataset such as the proposed football dataset. This means the model can generate good video from multi-views given one fixed trajectory. This proves the potential of the model for video generation from many different camera poses, which is promising in many fields such as 3D reconstruction.

\subsection{Limitation of Tora}
Tora \cite{zhang2024tora} is a recent trajectory-based diffusion model. We found Tora cannot understand object identity well, so it generates poor object interaction videos. As shown in Fig\,\ref{tora_problem}, objects in the videos mistakenly exchange trajectories when they are close to each other. However, for InTraGen, the trajectory remains stable not only when objects approach each other but also during collisions and overlaps, without confusion in the trajectory paths.

\section{More Implementation Details}
\subsection{Video Generation Process for InTraGen}
\begin{algorithm}[h] 
	\caption{Video Generation Process for InTraGen} 
	\label{alg::conjugateGradient}
	\begin{algorithmic}[1] 
		\Require 
		$(x_i,y_i)_{i:=1:N_0}$: trajectory (a list of coordinates). 
            \Require 
            $t_{text}$: text prompt to briefly describe the generation scene.
            \Require DiT model $\epsilon_{\theta}(.)$, VAE encoder $\mathcal{E}(.)$, VAE decoder $\mathcal{D}(.)$, text tokenizer/encoder $\mathcal{T}(.)$, Sampling Scheduler $\mathcal{S}(.)$
		\State \textbf{Initial} Noise initial latent $z_{1:N}^{T}$.
            \State $\triangleright$ Generate latent object ID maps and sparse Poses
            \State\,\,\, Calculate point differences $(dx_i,dy_i)=(x_i-x_{i-1},y_i-y_{i-1})_{i:=2:N_0}$
            \State\,\,\, Calculate sparse optical flow $(ox_i,oy_i)_{i:=2:N_0}=(ox_{i-1}+dx_i, oy_{i-1}+dy_i)$ and $(ox_1,oy_1)=(0, 0)$
            \State\,\,\, Calculate sparse poses $(sx_i,sy_i)_{i:=1:N_0}=\text{GaussianFilter}((ox_i,oy_i)_{i:=1:N_0})$
            \State\,\,\, Calculate object ID maps $(idx_i,idy_i)_{i:=1:N_0}=\text{CoordToColor}((x_i,y_i)_{i:=1:N_0})$
            \State\,\,\, Generate latent object ID and sparse poses: $ObjID=\mathcal{E}((idx_i,idy_i)_{i:=1:N_0})$ and $SpaPos=\mathcal{E}((sx_i,sy_i)_{i:=1:N_0})$
            \State $\triangleright$ Generate text conditions $c_{text} = \mathcal{T}(t_{text})$
            \For{t=T,...,1};
                \State $\hat{\epsilon} \leftarrow \epsilon_\theta\left(z_{1:N}^{t}, t, ObjID, SpaPos, c_{text}\right)$
                \State $z_{1:N}^{t-1}\leftarrow \mathcal{S}(z_t,\hat{\epsilon},t)$
            \EndFor
        \State Compute output video $v_{1:N_0}=\mathcal{D}(z_{1:N}^{0})$
	\end{algorithmic} 
 \label{algorithm}
\end{algorithm}

The video generation process of the proposed InTraGen is shown in Algorithm\,\ref{algorithm}. Given input trajectory $(x_i,y_i)_{i:=1:N_0}$ and text prompt $t_{text}$, we first calculate the latent object ID maps $ObjID$, sparse poses $SpaPos$, and text condition $c_{text}$. To obtain the sparse poses, we calculate the point difference as $(dx_i,dy_i)=(x_i-x_{i-1},y_i-y_{i-1})_{i:=2:N_0}$. Then, we use point difference to get the sparse optical flow, as $(ox_i,oy_i)_{i:=2:N_0}=(ox_{i-1}+dx_i, oy_{i-1}+dy_i)$. We further use sparse optical flow to calculate the sparse poses with Gaussian filter, as $(sx_i,sy_i)_{i:=1:N_0}=\text{GaussianFilter}((ox_i,oy_i)_{i:=1:N_0})$. The object ID is calculated by $(idx_i,idy_i)_{i:=1:N_0}=\text{CoordToColor}((x_i,y_i)_{i:=1:N_0})$, where $\text{CoordToColor}$ draw the colorful point for different objects to make them recognizable by the model. The text condition is generated by text tokenizer and text encoder, as $c_{text} = \mathcal{T}(t_{text})$. In the diffusion sampling process, the DiT model $\epsilon_{\theta}(.)$ is conditioned by object ID maps, sparse poses, and text condition to make the generated video follow the trajectory with realistic object interaction.

\subsection{More Explanation for the Select Metrics}
In the paper, we totally use 5 metrics including matching trajectory evaluation metric (MTEM), Peak Signal-to-Noise Ratio (PSNR), Structural Similarity Index Measure (SSIM), Learned Perceptual Image Patch Similarity (LPIPS), and Fréchet Image Distance (FID). The MTEM evaluates whether the generated objects follow the prompt trajectories. 
We apply PSNR in our task because the trajectory and text conditions in quantitative comparison are from their corresponding original videos in our proposed four datasets. Therefore, in perfect cases, the model should generate videos with good pixel alignment with the original videos. Note that it is not necessary to prompt our model with trajectory and text conditions from original videos, and we did this only for fair quantitative comparison. SSIM is used to calculate the perceptual quality of the generated videos. LPIPS is also applied to measure the high-level visual quality. FID is calculated at the frame level, and we average the frame-level FIDs to get the final score. The FID is used to compare the feature distribution between the real videos and generated videos. Some of these metrics may not be suitable for the unconditional settings. Nevertheless, for the conditional generation, besides the proposed metric MTEM, other metrics are commonly used in video generative settings~\cite{voleti2022mcvd,zhang2024extdm}. Note that our method performs the multi-conditional video generation using both text and trajectories.

\section{Failing Cases and Limitations}
\begin{figure*}[h]
    \centering
    \includegraphics[width=\linewidth]{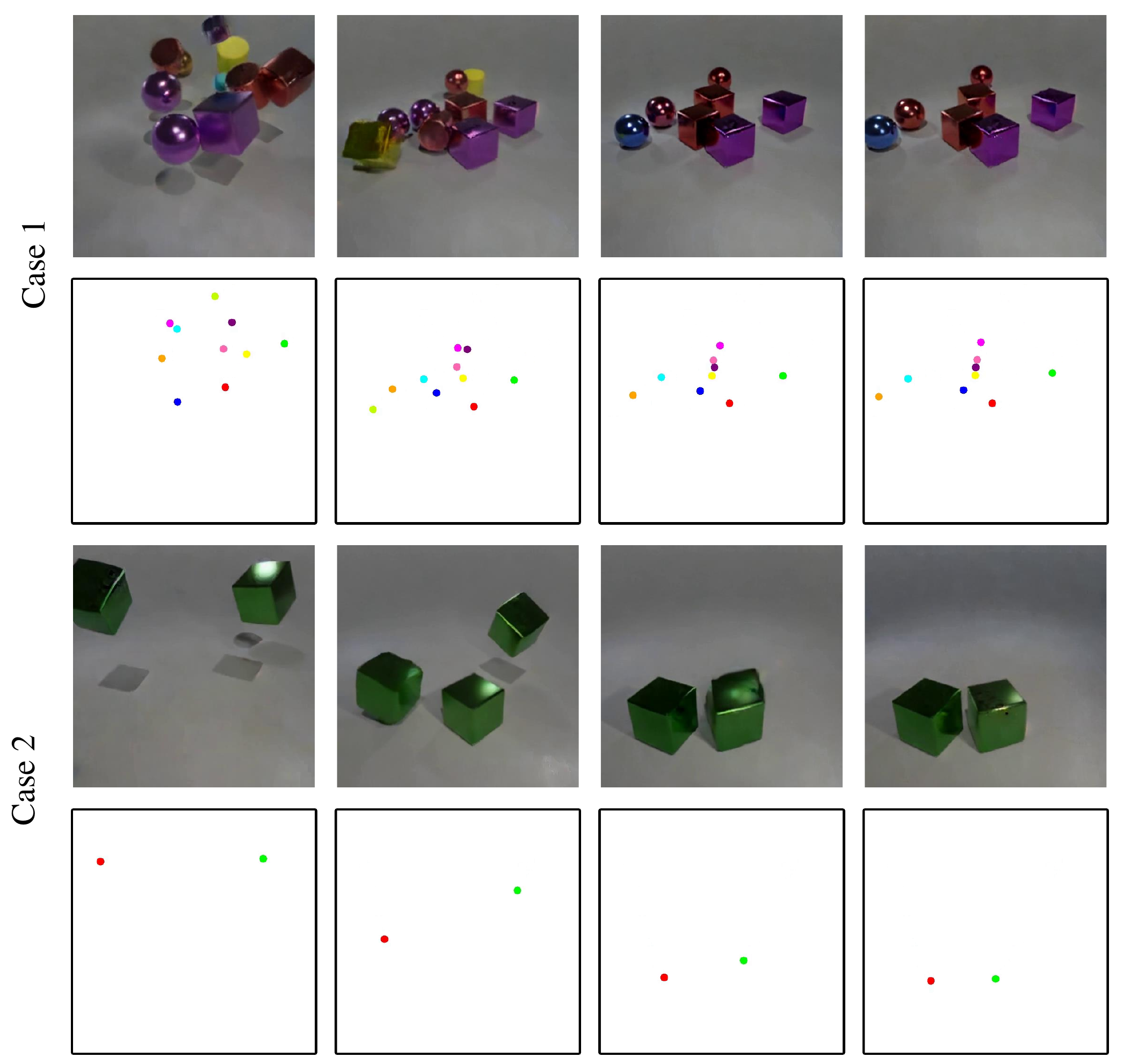}
    \caption{Failing cases. Case 1 - introducing too many objects close to each other makes it difficult for the model to distinguish between them due to the sparsity of the textual tokens and their inability to describe the scene in detail - this highly impacts the temporal consistency of tracked objects. For case 2, sometimes the model creates additional objects along the trajectory which merge into one as the video progresses.}
    \label{failing_cases}
\end{figure*}

Although the model achieves excellent performance in different datasets, it still have some limitations with falling cases happen during the video generation. Some examples can be found in Fig.\,\ref{failing_cases}. In Fig.\,\ref{failing_cases}, Case 1 introduces too many objects close to each other, which makes it difficult to distinguish between them due to the sparsity of the textual tokens and their inability to describe the scene in detail - this highly impacts the temporal consistency of tracked objects. For case 2, sometimes the model creates additional objects along the trajectory which merge into one as the video progresses.

\end{document}